%% file: root.tex
\documentclass[letterpaper, 10 pt, conference]{ieeeconf}  

\IEEEoverridecommandlockouts                              

\usepackage{algorithm}
\usepackage{algpseudocode}
\usepackage{xcolor}
\usepackage{booktabs}
\usepackage{multirow}
\usepackage{xspace}

\usepackage{cite}
\usepackage{amsmath}
\usepackage{amssymb}
\usepackage{amsfonts}
\usepackage[final]{graphicx}
\usepackage{textcomp}
\usepackage{etoolbox}
\usepackage{comment}
\usepackage{hyperref}
\usepackage{cleveref}
\usepackage{flushend}

\title{\LARGE \bf
Learning Robot Structure and Motion Embeddings\\
using Graph Neural Networks
}

\author{
J. Taery Kim$^{1}$, Jeongeun Park$^{2}$, Sungjoon Choi$^{2}$, Sehoon Ha$^{1}$
\thanks{$^{1}$
J. Taery Kim and Sehoon Ha are with Georgia Institute of Technology, Atlanta, GA, 30308, USA
$ \{ \tt \footnotesize taerykim, \, sehoonha \} $@gatech.edu
}
\thanks{$^{2}$ 
Jeongeun Park and Sungjoon Choi are with the Department of Artificial Intelligence, Korea University, Seoul, Korea
$ \{ \tt \footnotesize baro0906, \, sungjoon-choi \} $@korea.ac.kr
}%
}

\begin{document}

\maketitle
\thispagestyle{empty}
\pagestyle{empty}

\begin{abstract}
We propose a learning framework to find the representation of a robot's kinematic structure and motion embedding spaces using graph neural networks (GNN). Finding a compact and low-dimensional embedding space for complex phenomena is a key for understanding its behaviors, which may lead to a better learning performance, as we observed in other domains of images or languages. However, although numerous robotics applications deal with various types of data, the embedding of the generated data has been relatively less studied by roboticists. To this end, our work aims to learn embeddings for two types of robotic data: the robot's design structure, such as links, joints, and their relationships, and the motion data, such as kinematic joint positions. Our method exploits the tree structure of the robot to train appropriate embeddings to the given robot data. To avoid overfitting, we formulate multi-task learning to find a general representation of the embedding spaces. We evaluate the proposed learning method on a robot with a simple linear structure and visualize the learned embeddings using t-SNE. We also study a few design choices of the learning framework, such as network architectures and message passing schemes.
\end{abstract}

\input{defs}

\input{1_intro}
\input{2_related}
\input{3_method}

\input{4_experiment}

\input{5_conclusion}


\bibliographystyle{IEEEtran}
\bibliography{bib}


\addtolength{\textheight}{-12cm}   


\end{document}

%% file: defs.tex

\newcommand{\cmt}[1]{}
\newcommand{\updated}[1]{\textcolor{black}{{#1}}}
\newcommand{\taery}[1]{\textcolor{blue}{{Taery: #1}}}
\newcommand{\sehoon}[1]{\textcolor{red}{{Sehoon: #1}}} 
\newcommand{\sungjoon}[1]{\textcolor{magenta}{{Sungjoon: #1}}} 
\newcommand{\jungeun}[1]{\textcolor{cyan}{{Jungeun: #1}}} 
\newcommand{\newtext}[1]{#1}
\newcommand{\eqnref}[1]{Equation~(\ref{eq:#1})}
\newcommand{\figref}[1]{Figure~\ref{fig:#1}}
\renewcommand{\algref}[1]{Algorithm~\ref{alg:#1}}
\newcommand{\tabref}[1]{Table~\ref{tab:#1}}
\newcommand{\secref}[1]{Section~\ref{sec:#1}}

\long\def\ignorethis#1{}

\newcommand{\etal}{{\em{et~al.}\ }}
\newcommand{\eg}{e.g.\ }
\newcommand{\ie}{i.e.\ }

\newcommand{\figtodo}[1]{\framebox[0.8\columnwidth]{\rule{0pt}{1in}#1}}



\newcommand{\pdd}[3]{\ensuremath{\frac{\partial^2{#1}}{\partial{#2}\,\partial{#3}}}}

\newcommand{\mat}[1]{\ensuremath{\mathbf{#1}}}
\newcommand{\set}[1]{\ensuremath{\mathcal{#1}}}

\newcommand{\vc}[1]{\ensuremath{\mathbf{#1}}}
\newcommand{\vEndEff}{\ensuremath{\vc{d}}}
\newcommand{\vRelMove}{\ensuremath{\vc{r}}}
\newcommand{\sSet}{\ensuremath{S}}

\newcommand{\vControl}{\ensuremath{\vc{u}}}
\newcommand{\vPoint}{\ensuremath{\vc{p}}}
\newcommand{\sSpringCoef}{{\ensuremath{k_{s}}}}
\newcommand{\sDamperCoef}{{\ensuremath{k_{d}}}}
\newcommand{\vHandle}{\ensuremath{\vc{h}}}
\newcommand{\vForce}{\ensuremath{\vc{f}}}

\newcommand{\mTransChain}{\ensuremath{\vc{W}}}
\newcommand{\mRotateTrans}{\ensuremath{\vc{R}}}
\newcommand{\sJoint}{\ensuremath{q}}
\newcommand{\vJoint}{\ensuremath{\vc{q}}}
\newcommand{\mJoint}{\ensuremath{\vc{Q}}}
\newcommand{\mMass}{\ensuremath{\vc{M}}}
\newcommand{\sMass}{\ensuremath{{m}}}
\newcommand{\vGravity}{\ensuremath{\vc{g}}}
\newcommand{\vConstr}{\ensuremath{\vc{C}}}
\newcommand{\sConstr}{\ensuremath{C}}
\newcommand{\vCOM}{\ensuremath{\vc{x}}}
\newcommand{\sGeneralForce}[1]{\ensuremath{Q_{#1}}}
\newcommand{\vStateVar}{\ensuremath{\vc{y}}}
\newcommand{\vControlVar}{\ensuremath{\vc{u}}}
\newcommand{\tr}[1]{\ensuremath{\mathrm{tr}\left(#1\right)}}

%
%

\renewcommand{\choose}[2]{\ensuremath{\left(\begin{array}{c} #1 \\ #2 \end{array} \right )}}

\newcommand{\gauss}[3]{\ensuremath{\mathcal{N}(#1 | #2 ; #3)}}

\newcommand{\pctab}{\hspace{0.2in}}
\newenvironment{pseudocode} {\begin{center} \begin{minipage}{\textwidth}
                             \normalsize \vspace{-2\baselineskip} \begin{tabbing}
                             \pctab \= \pctab \= \pctab \= \pctab \=
                             \pctab \= \pctab \= \pctab \= \pctab \= \\}
                            {\end{tabbing} \vspace{-2\baselineskip}
                             \end{minipage} \end{center}}
\newenvironment{items}      {\begin{list}{$\bullet$}
                              {\setlength{\partopsep}{\parskip}
                                \setlength{\parsep}{\parskip}
                                \setlength{\topsep}{0pt}
                                \setlength{\itemsep}{0pt}
                                \settowidth{\labelwidth}{$\bullet$}
                                \setlength{\labelsep}{1ex}
                                \setlength{\leftmargin}{\labelwidth}
                                \addtolength{\leftmargin}{\labelsep}
                                }
                              }
                            {\end{list}}
\newcommand{\newfun}[3]{\noindent\vspace{0pt}\fbox{\begin{minipage}{3.3truein}\vspace{#1}~ {#3}~\vspace{12pt}\end{minipage}}\vspace{#2}}

\newcommand{\key}{\textbf}
\newcommand{\fun}{\textsc}



%% file: 1_intro.tex
\section{Introduction}

The recent success of deep learning heavily relies on the representation learning of input data. However, most representation learning methods have focused on structured data, such as images or texts~\cite{dong2017metapath2vec,kolesnikov2019revisiting,chen2021exploring}. On the other hand, researchers have made less attention to understand various robotics data, such as robots' structures, motions, controls, and sensory observations, where their structures remain unveiled. In this paper, we focus on the problem of understanding the robot's data by finding its low-dimensional embedding space.



Given the fact that the robot's structure can be represented as a graph, it seems to be a natural choice to leverage Graph neural networks (GNNs) to represent the robot's data.
Particularly, we employ the framework of message passing neural networks (MPNNs) proposed by Gilmer et al.~\cite{gilmer2017neural} as the network architecture, which extends the message passing schemes between nodes. This scheme has been applied to learn a variety of graph-structured data, from a single large graph (\textit{e.g.}, social network) to multiple small graphs (\textit{e.g.}, molecules). While the former data typically involves the per-node or per-edge data, the latter focuses on graph-level tasks. Inspired by the success of molecule works~\cite{wu2018moleculenet,duvenaud2015convolutional,mansimov2019molecular}, we also formulate the learning with graph-level tasks that involve the data generated from the robot's entire structure.

In addition, we take an approach of multi-task learning (MTL) when we learn embeddings. This is a well-known approach for learning a shared representation that is agnostic to the selection of tasks while avoiding overfitting. In particular, we set two related tasks in this work: solving forward kinematics (FK) and inverse kinematics (IK) while sharing robot structure representation. Because the robot's morphologies and motions are related to each other, we expect that this simultaneous learning of FK and IK allows us to find general embeddings more efficiently.

We evaluate the proposed learning framework on a simple 2D robot with a linear structure. We generate a large size of data while varying the number of joints and link lengths, and augment the structure data set with pose data. Then we process the given multi-dimensional data to find a compact fixed eight-dimensional 
embedding space. We visualize the learned embedding space using t-SNE and compare it with the embedding of the multilayer perception (MLP), which gives us different insights into the robot design and motion spaces.

%% file: 2_related.tex

\begin{figure*}[ht]
    \centering
    \includegraphics[width=0.9\linewidth]{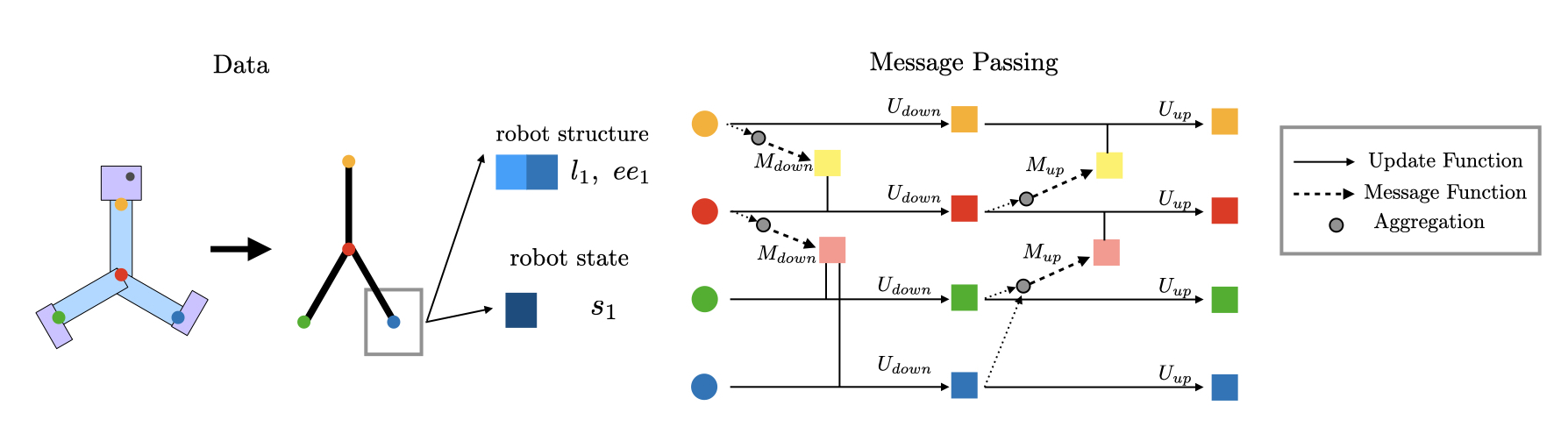}
    \vspace*{-0.3cm}
    \caption{Tree message passing overview. 
        (a) Robot structure converted to tree-structured data with nodes and edges. Robotic data such as joint angle $s$ and link length $l$ are assigned to node or edge features.
        (b) Two paths in message passing: \textit{downward passing} to spread the information from parent to child nodes, then \textit{upward passing} to aggregate the information at the root node. $M$, $A$, and $U$ denote the message generation, message aggregation, and hidden state update function, respectively, for each direction ${down}$ and ${up}$.
        }
    \label{fig:tree-msg-passing}
\end{figure*}

\section{Related Work}

In this work, we focus on the problem of embedding structural features of various robots through message passing neural networks~\cite{gilmer2017neural}. 
Our goal of finding the robot design embedding differs from other existing tasks~\cite{reily2020representing}, which do not consider any kinematic constraints. This section will review the prior work that leverages GNN for robotics and then go over more recent works on multi-task learning.

\subsection{Graph Neural Networks in Robotics}
The existing works on a single robot with a graph neural network can be categorized into two tasks, design optimization, and locomotion task.
The problem of design optimization, which aims to find the optimal design that can maximize the performance, can have a more structured search by obtaining embedded design space. 
Instead of manually constructing the manifolds~\cite{ha2017joint,ha2018computational}, the recent advances in GNN allow us to learn them from the data automatically. 
RoboGrammar~\cite{zhao2020robogrammar} aims to solve the problem of robot design in a terrain environment, which exploits GNN for estimating design value functions to efficiently search the robot design space defined by the predefined grammar. Xu et al.~\cite{xu2021multi} expand RoboGrammar to a multi-objective task, which uses graph heuristic search (GHS) iteratively to sample different directions of the objective spaces while learning a GNN-represented heuristic function shared across these directions. 
Instead of learning value functions, Wang et al.~\cite{wang2019neural} propose a technique to optimize the robot design using evolution strategy algorithms while evaluating designs using GNN based controllers.

In addition, a number of works~\cite{wang2018nervenet,huang2020one,whitman2021learning} seek to obtain a generalizable control policy without the notion of design optimization, although they are two related problems. 
NerveNet~\cite{wang2018nervenet} seeks to train generalized locomotion skills using GNN that can be transferred to the robot with a different number of joints or even with broken actuators in a zero-shot manner. Modular RL~\cite{huang2020one} introduces Shared Modular Policies (SMP), a structure agnostic optimal locomotion policy based on each module by utilizing two-way message passing between them. 
Whitman et al.~\cite{whitman2021learning} also propose an algorithm to find a kinematic chain agnostic control policy by leveraging both a message-passing network on the robot structure and a model-based reinforcement learning algorithm.

\subsection{Multi-task Representation Learning.}
In this work, we perform multi-task learning (i.e., forward kinematics and inverse kinematics)to find the structural embedding of robot designs. We started with the Shared Trunk architecture~\cite{zhang2014facial,dai2016instance} that utilizes a global feature extractor shared for all tasks with designated output heads. Prediction distillation~\cite{xu2018pad,vandenhende2020mti} extends this idea based on the intuition that the answer for one task will help in learning another task. Similarly, we can formulate an auxiliary task to boost the learning of general representation without decreasing the performance of the original task. For instance, Liebe et al.~\cite{liebel2018auxiliary} targets vision-based road scene understanding by defining auxiliary tasks as unrelated tasks but has similar features as the main tasks. Furthermore, one can use multiple input domains (e.g., image and text) with each feature extractor, and this structure is called Multi-Modal architecture. Giannone et al.~\cite{giannone2019learning} presents architecture composed of two encoders for RGB and depth images and concatenate them to obtain common representation for both semantic segmentation and depth completion. Our proposed method belongs to Multi-Modal architecture that is similar to Giannone et al., where we simultaneously learn structure and pose embeddings by solving multiple tasks.
 

%% file: 3_method.tex
\section{Learning Robot Embedding with GNN} \label{sec-method}

A well-formed robot design embedding has the potential to solve structure-relevant problems such as design optimization or robot-agnostic tasks.
Our method encodes robotic data into two separated spaces: structure embedding and pose embedding.
The pose embedding is decoupled from the structure embedding, so that structure embedding targets particularly for invariant information which is specific to the robot structure. 
Instead, the pose embedding helps to capture kinematics information, which fundamentally involves robots' physical feasibility, robotic properties, or skills expected to learn through the embeddings.
To this end, we use graph neural networks to embed diverse robotic structures with non-fixed dimensional robotic data into a fixed dimensional space.

In this section, we suggest a method to exploit a tree structure on learning representations of robotic data. 
Then we describe how our method employs message passing neural networks and creates the embedding space through multi-task representation learning.

\begin{figure}[t]
    \centering
    \includegraphics[width=1\linewidth]{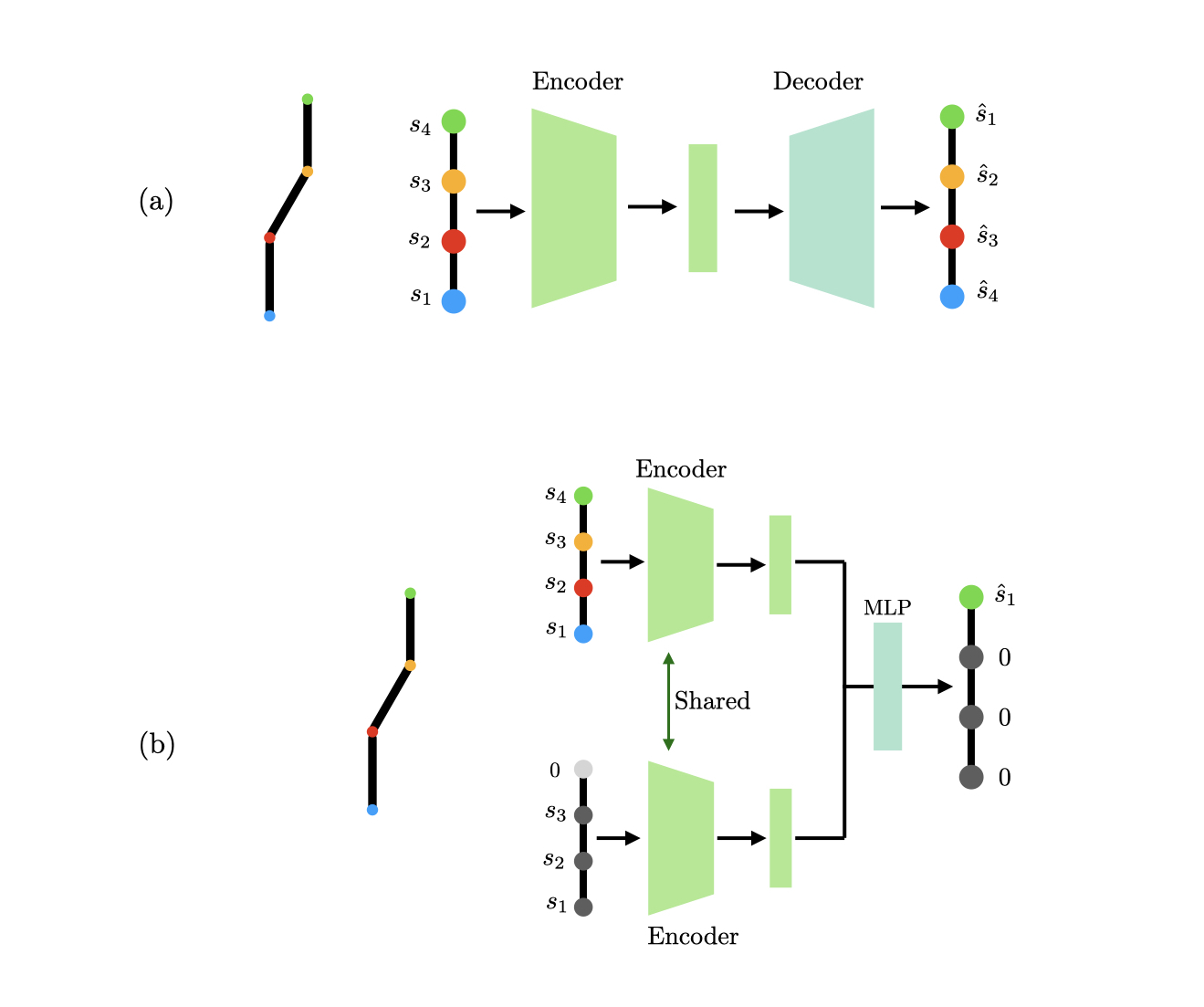}
    \vspace{-0.8cm}
    \caption{Two reconstruction tasks to pretrain embedding space: (a) Encoder-decoder (ED) architecture reconstructs the features for all the nodes; (b) Fill-in-the-blank (FB) architecture randomly masks out single node to be reconstructed. Learned embedding spaces are compared in Fig.~\ref{fig:latent-vis-reconstruction}.
    }
    \label{fig:recon}
\end{figure}

\begin{figure}
    \centering
    \includegraphics[width=1\linewidth]{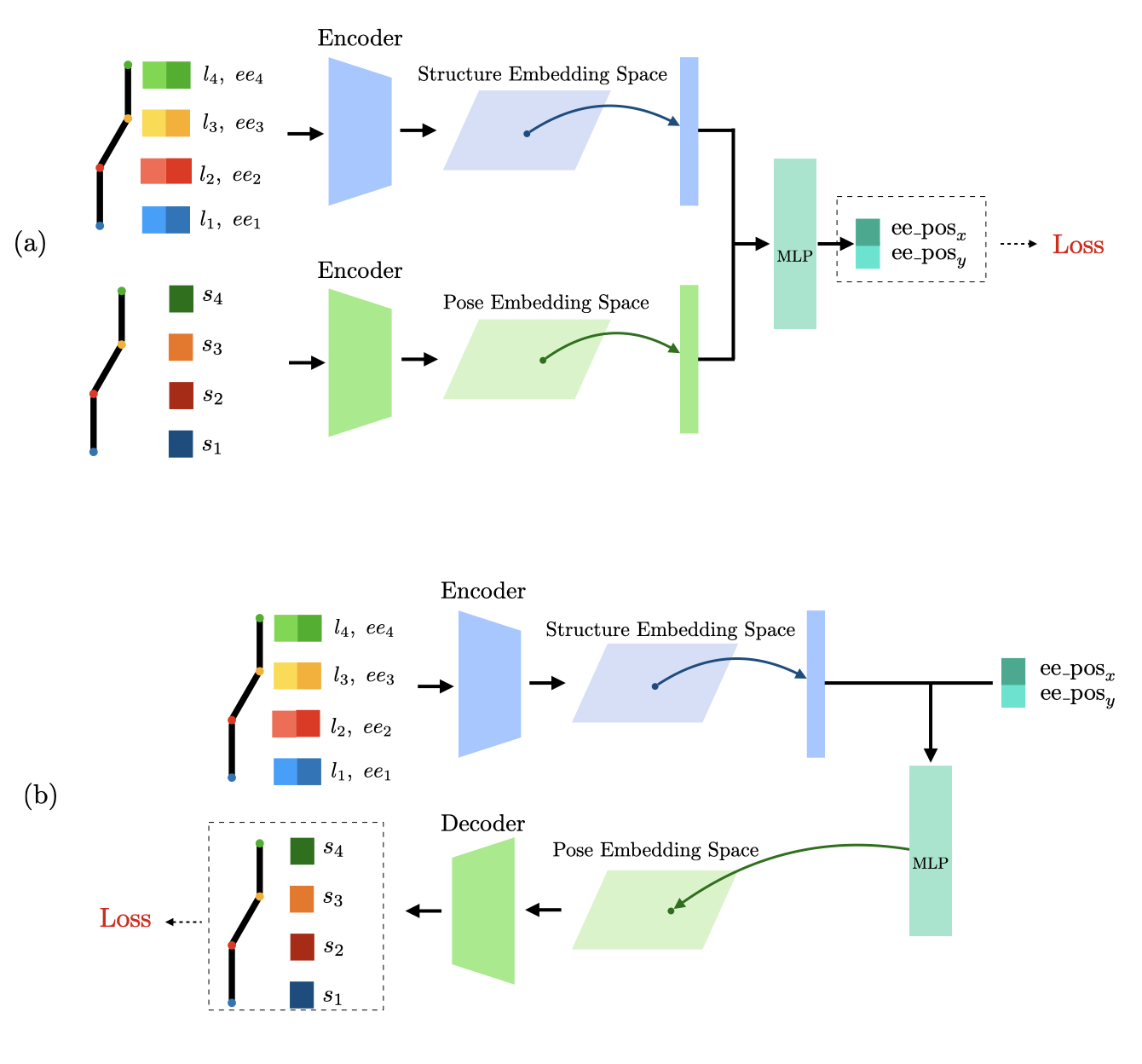}
    \vspace{-0.8cm}
    \caption{Two tasks solved jointly as multi-task learning: (a) forward kinematics, FK; and (b) inverse kinematics, IK. The tasks share structure embedding space and pose embedding space while training.
    FK predicts the end-effector position from the structure  and pose embeddings, while IK estimates the pose from the structure embedding and end-effector position.
    }
    \label{fig:mtl}
\end{figure}

\subsection{Tree Message Passing}
We can model a robot's structure as a tree by converting joints to nodes and links to edges while assuming no loops as shown in Fig.~\ref{fig:tree-msg-passing}a.
Then a leaf node represents a null joint that makes the connected edge an end-effector.
We focus on this tree structure of the robots and exploit it to learn compact but meaningful structural or pose embeddings. 

Since a tree is a special case of graphs, our method starts with the message passing strategy, generating messages at nodes and passing them via edges.
Unlike undirected graphs, trees have directional relationships between parent and child nodes that define the hierarchy of nodes. 
We explicitly use this hierarchy on the message passing scheme.

Fig.~\ref{fig:tree-msg-passing}b describes how the messages are generated and passed throughout the tree structure. 
Each message passing step updates the hidden state $h_r$ of receiver node $r$ with message function $M$, aggregation function $A$, and update function $U$.
Let us assume a set of $n$ sender nodes $\mathcal{S} \in [s_1, s_2, \cdots, s_n]$ that need to send their message to the receiver node $r$.
Then each node $s \in \mathcal{S}$ will generate its own message using the function $M$:
\begin{equation}
m_{s}^t = M_{dir}(h_s^t, e_{(s,r)}),
\end{equation}
where ${dir}$ denotes the direction of the message passing, and $e_{s,r}$ denotes the feature of the edge $(s,r)$.
Either parent or child node can be the sender based on the direction of the message passing.
Then, the messages from the senders are aggregated by the aggregation function \updated{$A_{dir}$} as
\begin{equation}
m^t = A_{dir}(m_{s_1}^t, m_{s_2}^t, \cdots,  m_{s_n}^t).
\end{equation}
Finally, the hidden state of the receiver node $r$ is updated as
\begin{equation}
h_r^{t} = U_{dir}(h_r^{t-1}, m^{t})
\end{equation}
where $t$ denotes the time step that the hidden state be updated.

Within the tree structure, we define message passing in two directions, from root to leaf or leaf to root, which will be referred to as \emph{downward passing} and \emph{upward passing}, respectively. 
For the downward passing, each node is updated only based on its parent node, thus an identity function is used as an aggregation function $A$.
This passing can also be viewed as spreading a parent node's information to child nodes.
Contrarily, upward message passing has to collect the information from multiple child nodes using an aggregation function.
Typical choices of an aggregation function include mean, pooling, and recurrent neural networks.

\subsection{Pretraining of Encoders}\label{subsec:recon}
Ahead of multi-task learning, we pre-train both structure and pose encoders to facilitate the whole learning process.
This is because we observe that the proposed multi-task learning is difficult to learn from scratch.
For this purpose, we design new self-reconstructing tasks that encode node features and decode to reconstruct the given inputs.
Because each node feature may consist of more than one type of information, we also consider this task MTL.

The straightforward formulation is to reconstruct all the node features (Fig.~\ref{fig:recon}a) in the encoder-decoder (ED) architecture.
Both encoder and decoder use the tree message passing networks.
Because each data may have a different tree structure, we stick to a pre-order traversal, starting from the root node.
However, the ambiguity is still not fully considered, as identical robot structures may have different tree representations.

We also investigate another method for pretraining, which is to reconstruct the missing node's features from the given other nodes' features.
This approach is inspired by fill-in-the-blank tasks in other domains, such as text infilling in natural language processing~\cite{yu2015visual, joshi2020spanbert, devlin2018bert} and image inpainting in computer vision~\cite{bertalmio2000image}.
We leave one node out randomly and let the networks only predict the blanked information rather than reconstructing all the node features.
Both the whole graph and blank graph are encoded with the same tree encoder, and then an MLP predicts the blanked information as in Fig.~\ref{fig:recon}b, called a fill-in-the-blank (FB) architecture.
We expect that this method would capture even minor differences in each node features and also relieve the ambiguity compared to ED reconstruction.
For simplicity, we concatenate edge features into its child node features, and treat everything as node features.

\subsection{Multi-task Representation Learning on Kinematics}
We suggest learning an embedding space of robot through multi-task representation learning (MTL) as illustrated in Fig.~\ref{fig:mtl} to learn the space that is generalizable across multiple tasks. 
Because our goal is to make the embedding space of the robot's kinematics, we select two popular tasks: forward kinematics and inverse kinematics.

As shown in Fig.~\ref{fig:mtl}, the tasks are sharing structure and pose embedding space. 
We separate structure-specific data to create its own embedding space.
For the structure embedding, we use link lengths $l$ and binary flags $ee$ to note whether the node is an end-effector as node features.
For the pose embedding, we use joint positions $s$ as the node features, which vary in a single robot structure.

The first task, forward kinematics (FK, Fig.~\ref{fig:mtl}a), is to find the end-effector position in Cartesian space from the given joint angles.
First, we obtain a structure embedding of a robot using a structure-tree encoder while computing a pose embedding from joint angles using a pose-tree encoder. 
Then, both the structure and  pose embeddings are concatenated and fed to a multilayer perceptron (MLP) to predict the end-effector position.

On the other hand, the inverse kinematics task (IK, Fig.~\ref{fig:mtl}b) predicts joint angles from the given position of the end-effector.
Note that the IK task naturally has ambiguity with multiple solutions while the FK task predicts the unique end-effector position.
The structure embedding is obtained through the aforementioned structure-tree encoder and then combined with the given end-effector position.
Then an MLP layer maps the combined information to the pose-tree embedding space to reconstruct the pose-tree structure, which holds the joint angle configuration.

We weight each loss based on initial loss to prevent the learning dominated by one task and linearly combine as
\begin{equation}
\mathcal{L}_{total} = w_\text{FK}\mathcal{L}_\text{FK} + w_\text{IK}\mathcal{L}_\text{IK}.
\end{equation}
For the following experiments, the loss is computed using the mean absolute error between the targeted and predicted positions, and the weights $w_\text{FK}$ and $w_\text{IK}$ are set as 5.0 and 0.5, respectively.

%% file: 4_experiment.tex
\begin{figure}
    \centering
    \includegraphics[width=1\linewidth]{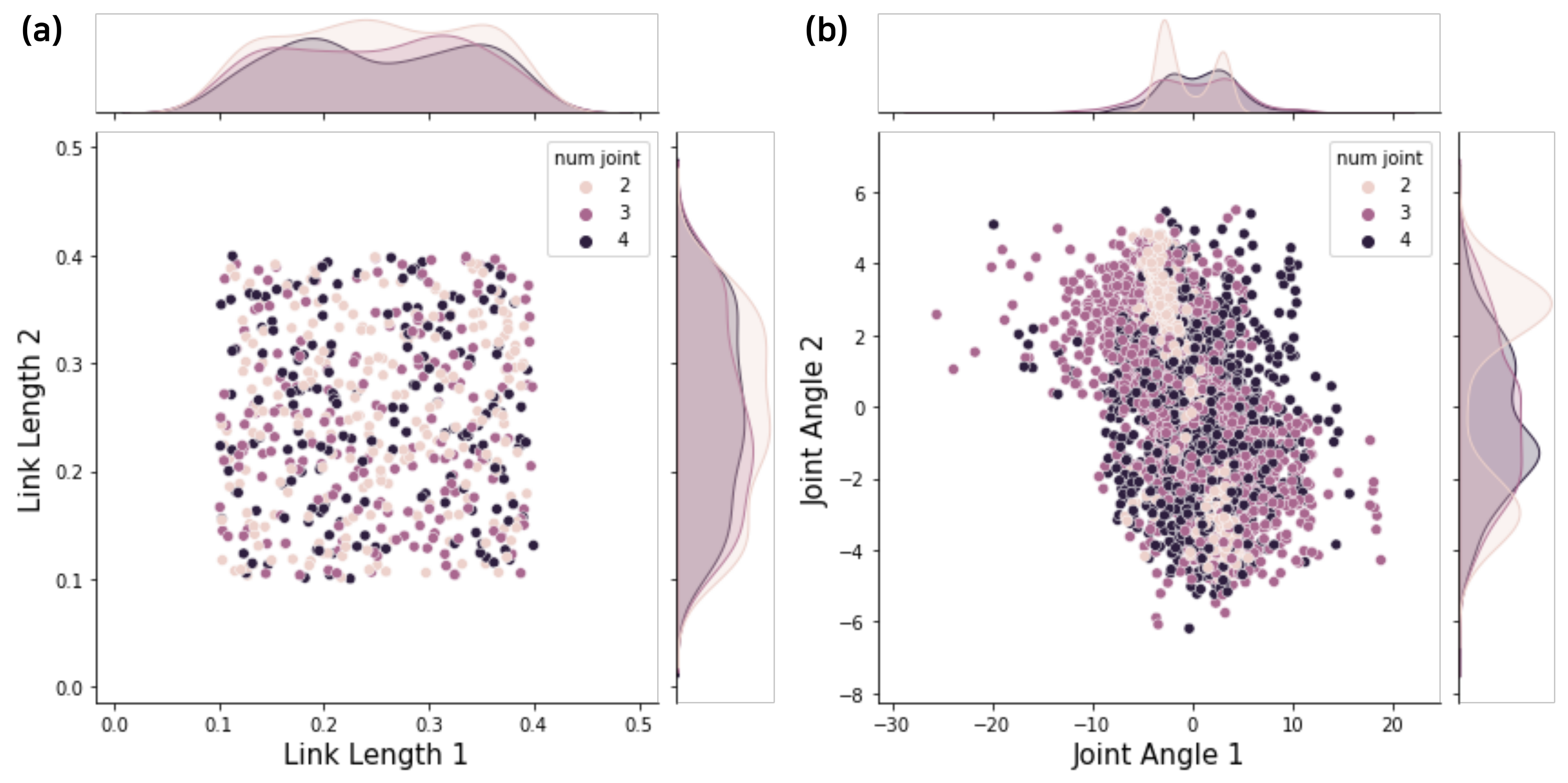}
    \vspace{-0.3cm}
    \caption{Data distribution of the first and second node features, labeled with the number of joints of each robot structure. (a) Structure data, link lengths, sampled from a uniform distribution. (b) Pose data, joint angles, computed values from random end-effector positions sampled from a uniform distribution.
    }
    \label{fig:data-distribution}
\end{figure}

\begin{figure}
    \centering
    \includegraphics[width=0.9\linewidth]{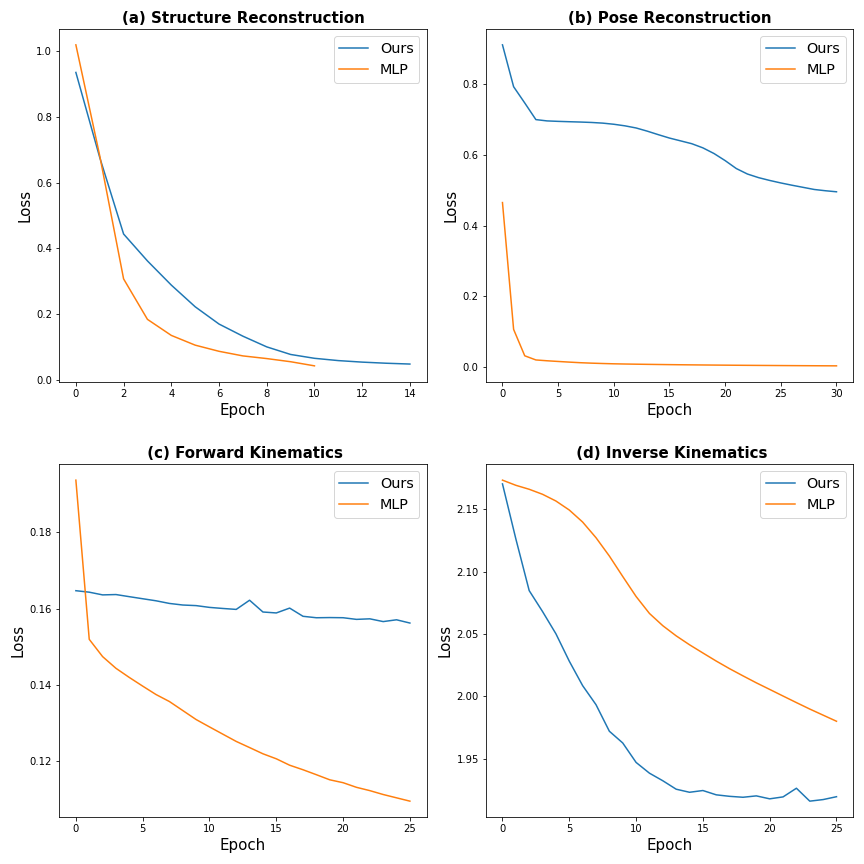}
    \vspace{-0.2cm}
    \caption{Learning curves compared with a MLP model. \textbf{Top:} pretraining of (a) structure and (b) pose embeddings.  \textbf{Bottom:} multi-task learning tasks (c) forward kinematics and (d) inverse kinematics.
    }
    \label{fig:learningcurve}
\end{figure}


\begin{figure*}
    \centering
    \includegraphics[width=1\linewidth]{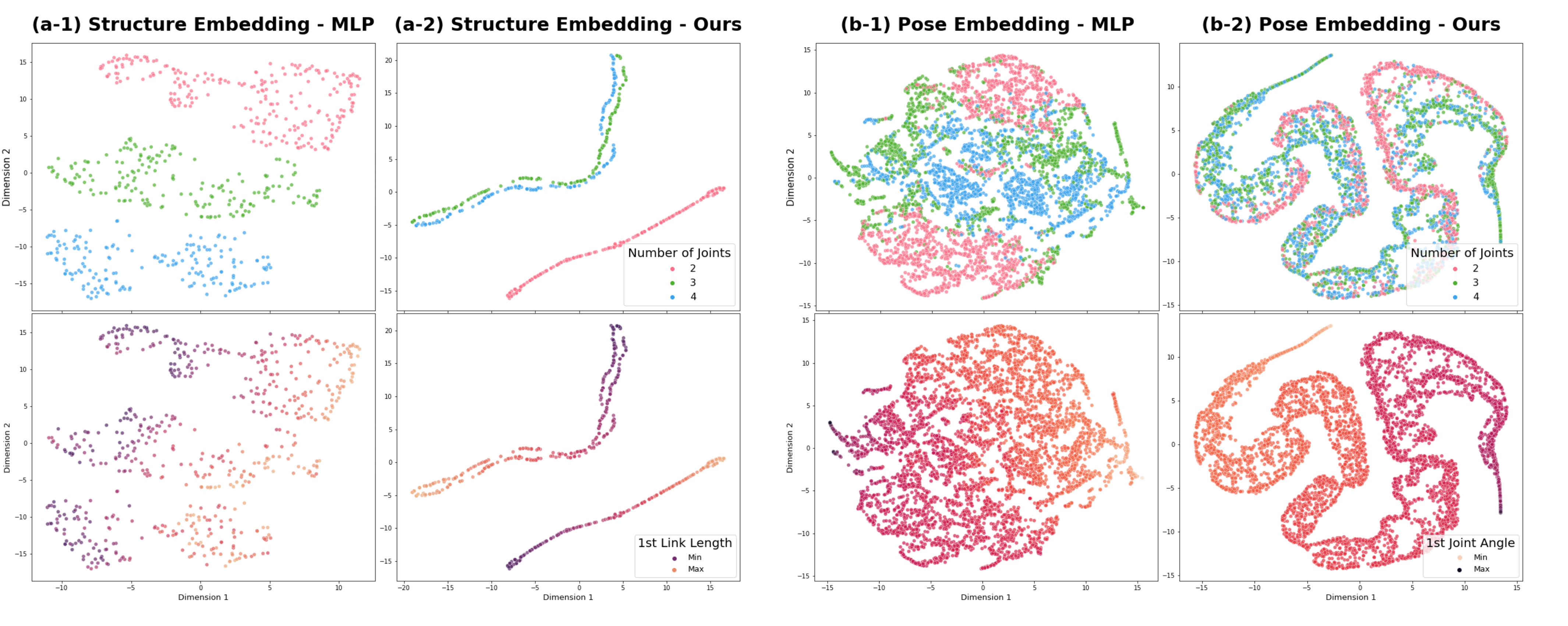}
    \vspace*{-0.9cm}
    \caption{
    Embedding space visualizations using t-SNE:
    (a) Structure embedding (b) Pose embedding.
    We plot the same embedding with two different color schemes, where they denote the number of joints in the upper row and the first link length in the bottom row.
    }
    \label{fig:latent-vis}
\end{figure*}

\section{Experiments}
This section presents experimental results of validating our approach.
We first describe the experimental setup, including data generation and implementation detail.
Then, we present the results on learning robot structural and pose embeddings and visualize the learned latent space to discover what is emphasized in the learning and how the message passing affects the learning.
In this pilot study, we focused on a simple robot with a linear structure.

\subsection{Data Generation}
We built our data set using the Reacher environment in OpenAI Gym~\cite{brockman2016openai}, which is a well-known benchmark in reinforcement learning.
A reacher is a robot arm with one end-effector, so we can represent the robot's structure with a tree with one child node for each parent node.
We vary the number of joints and the length of each link to generate various robot structures while changing the poses of each robot to learn kinematics.
As a result, the generated dataset consists of two distinct types of graphs: structure data and pose data.
Both are in tree structure holding node features and edge features, but for simplicity, we assign edge features to the child node's node features in this study as aforementioned.

In the robot structure data, the joint number varies from two to four, and the link lengths were randomly sampled from a uniform distribution $\mathcal{U}(0.1, 0.4)$. For each number of joints, we collected 1000 different robot structures. Each node feature includes link length and end-effector information, which are the unique attributes of each robot structure. 

On the other hand, the pose data is composed of joint configurations and end-effector positions, which are assigned as a node feature and a graph feature, respectively. We generate multiple pose data from one robot structure. We first sample 10,000 end-effector positions randomly from each robot structure, and joint configurations are calculated via inverse kinematics using the PyBullet library~\cite{coumans2019pybullet}. Then, we clean the data to filtrate any collision occurrence. 

Fig.~\ref{fig:data-distribution} shows the data distribution of (a) structure and (b) pose dataset, color labeled with respect to the number of nodes in the robot's structure. 
We randomly split the dataset into training and test sets with a ratio of 8:2.
The features of structure data are uniformly distributed as generated directly from sampling, whereas the pose data generation is biased a bit due to inverse kinematics computation.

\subsection{Implementation Details}
The structure and pose encoders are trained to encode the given input into a fixed-dimension vector, which is empirically set to 8.
The size of hidden layers in MLP is also set to 8.
The model is trained with the Adam optimizer~\cite{kingma2014adam} with an initial learning rate of 0.0001.
Each message passing with hidden state update repeated twice, both downward and upward so that every node knows the accumulated information, even of sibling nodes.

\subsection{Learning Curves}
In Fig.~\ref{fig:learningcurve}, we plot the progress of representation learning in terms of the loss values.
For comparison, an MLP encoder is trained to build the embedding space in the same dimension as our method does. 
When training MLP encoder, all inputs are expanded to the maximum number of joints with zero-filling for unused joints.
We draw various insights from the given learning curves.
First, both MLP and ours manage to lower the structure reconstruction loss to a similar scale.
However, our approach does not perform well in the pose reconstruction task and FK task in MTL.
We suspect that the reconstruction tasks are too easy for MLP because it can simply copy the inputs, which sharply drops the loss values to zero.
For the IK task, ours performs slightly better than the MLP encoder.
Originally, we expected that the proposed algorithm could effectively solve multiple tasks simultaneously with our learned embeddings, but the learning curves appear as pose embedding not learned well compared to the structure embedding.

\subsection{Embedding Space Visualization}
To explore the learned latent space, we plot the embedding vectors to two-dimensional space using t-SNE~\cite{van2008visualizing} as shown in Fig.~\ref{fig:latent-vis}.
The color in the upper row indicates the number of joints, while the bottom row figures are colored based on the first node's feature value, \textit{i.e.}, link length or joint angle.


\noindent \textbf{Structure embedding.}
Fig.~\ref{fig:latent-vis}a compares the structure embedding spaces learned with two architectures, our GNN with fill-in-blank architecture (column (a-1)) and the standard MLP (column (a-2)). We present both embeddings with two different colorizations based on the number of joints (upper) and the link length of the first node (lower). Both embeddings build clusters based on those two features, the number of joints and the link length of the first node, which affect the kinematic performance of the robot significantly, i.e., the reachable workspace of the robot. However, the overall shapes are quite different from each other: the MLP produces more dispersed clusters while our method creates two long-curved lines. We suspect that our method maintains a single cluster for three-linked and four-linked structures because both offer multiple solutions for a single IK problem. 

We argue that one embedding is not particularly better than the other. Rather, they just provide different perspectives for understanding the robot structure data. The MLP embedding space is not what we wanted to obtain because it does not give us any intuition for relating structures with different numbers of links. On the other hand, the embedding with our approach associates structures with three and four links, which can be reasonable in terms of their kinematic capabilities, such as reachability and singularity against IK.

\begin{figure}
    \centering
    \includegraphics[width=1\linewidth]{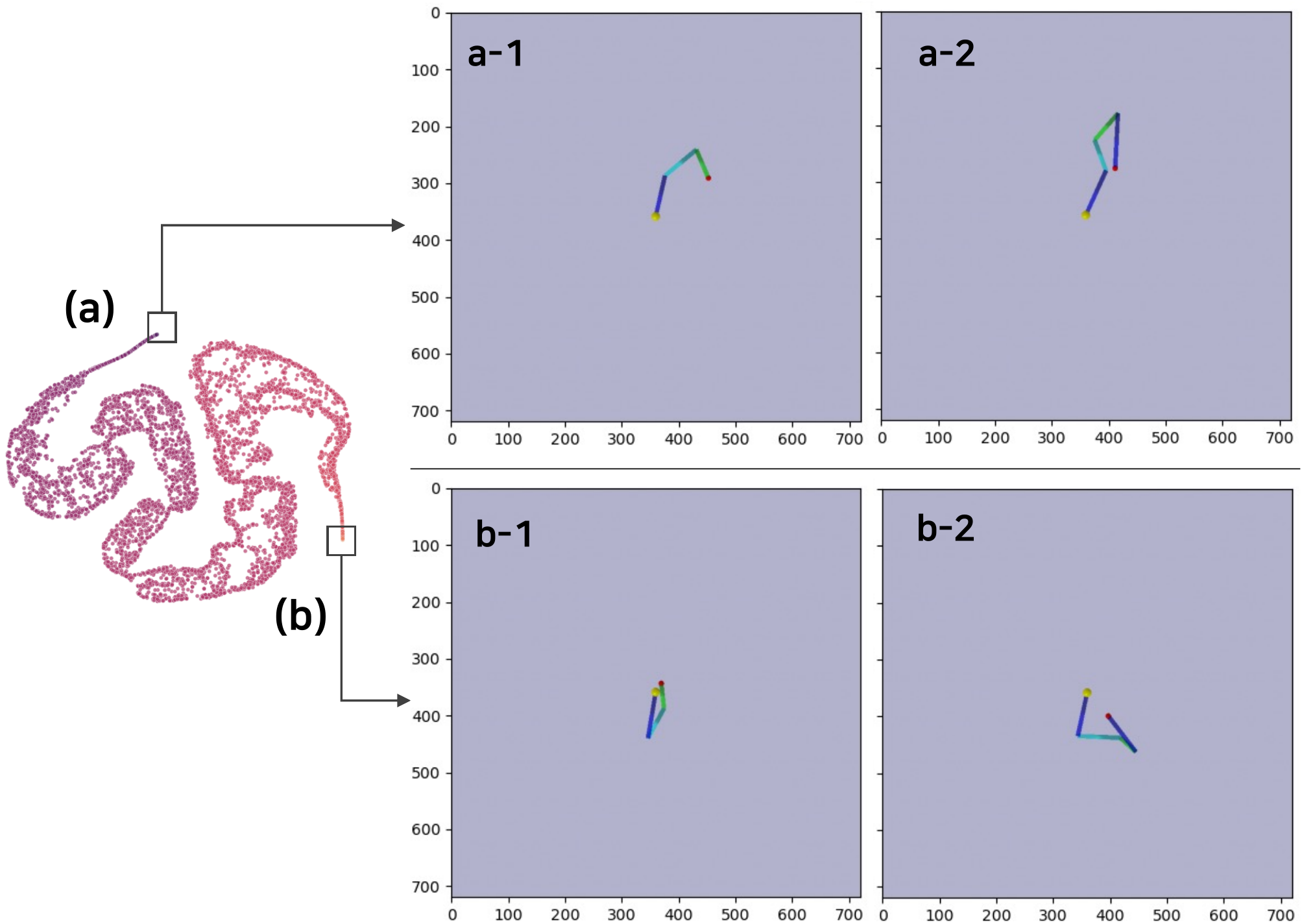}
    \caption{Sampled robot poses from pose embedding space. (a) and (b) are each end of the cluster. The pose embeddings are mostly determined by the first joint angle.
    }
    \label{fig:sampled-pose}
\end{figure}

\noindent \textbf{Pose embedding}
As shown in Fig.~\ref{fig:latent-vis}b, pose embeddings are in a single cluster and correlate to the node feature (i.e., joint angle), not the number of joints.
The embeddings of MLP (column (b-1)) are distributed all over the space without any recognizable shape, but ours are connected in a lengthy cluster with sharp ends (column (b-2)).
Fig.~\ref{fig:sampled-pose} shows the actual robot poses sampled from each end of the pose embedding space learned with our method. Left column robots have three joints, while right column robots have four joints.
Similar to our observation from the latent space visualization, the embedding is mostly correlated to the first joint angle that determines the range of end-effector position.

\subsection{Reconstruction Strategies Comparison}
\begin{figure}
    \vspace{0.1cm}
    \centering
    \includegraphics[width=1\linewidth]{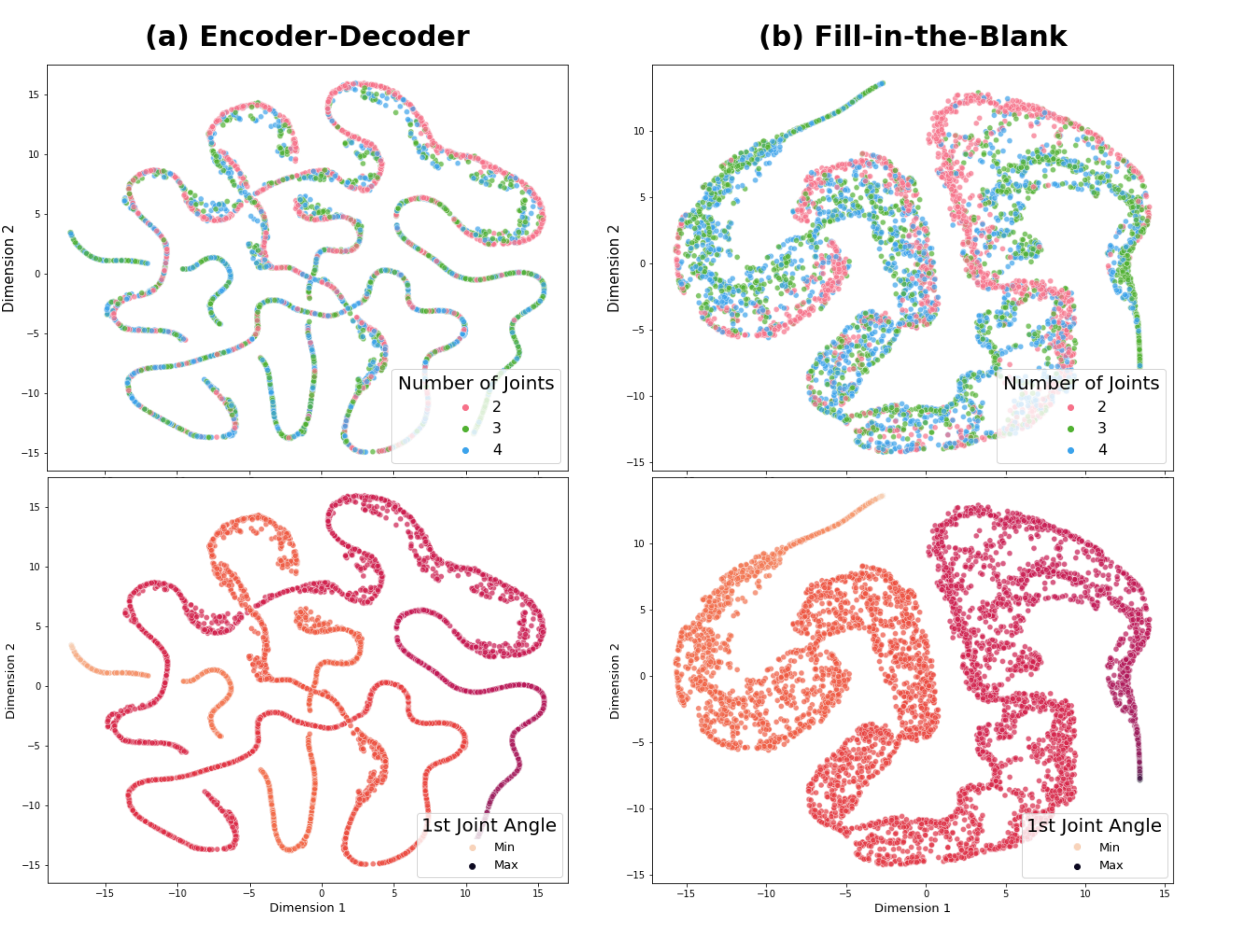}
    \vspace{-0.8cm}
    \caption{
    Comparison of reconstruction strategies. (a) Encoder-decoder (ED) (b) Fill-in-the-blank (FB).
    }
    \label{fig:latent-vis-reconstruction}
\end{figure}

In Fig.~\ref{fig:latent-vis-reconstruction}, we further compare two different reconstruction strategies discussed in Sec.~\ref{subsec:recon}.
ED reconstruction embeddings (Fig.~\ref{fig:latent-vis-reconstruction}a) create a long line connected all together with a bit of noise, but no specific correlation shown with neither the number of node nor node features.
Contrarily, the embeddings learned by FB reconstruction are in more stretched shapes but subject to the node features.
That is, FB reconstruction distinguishes each input in detail and be expected to facilitate further tuning in the embedding space.

\begin{figure}
    \centering
    \includegraphics[width=1\linewidth]{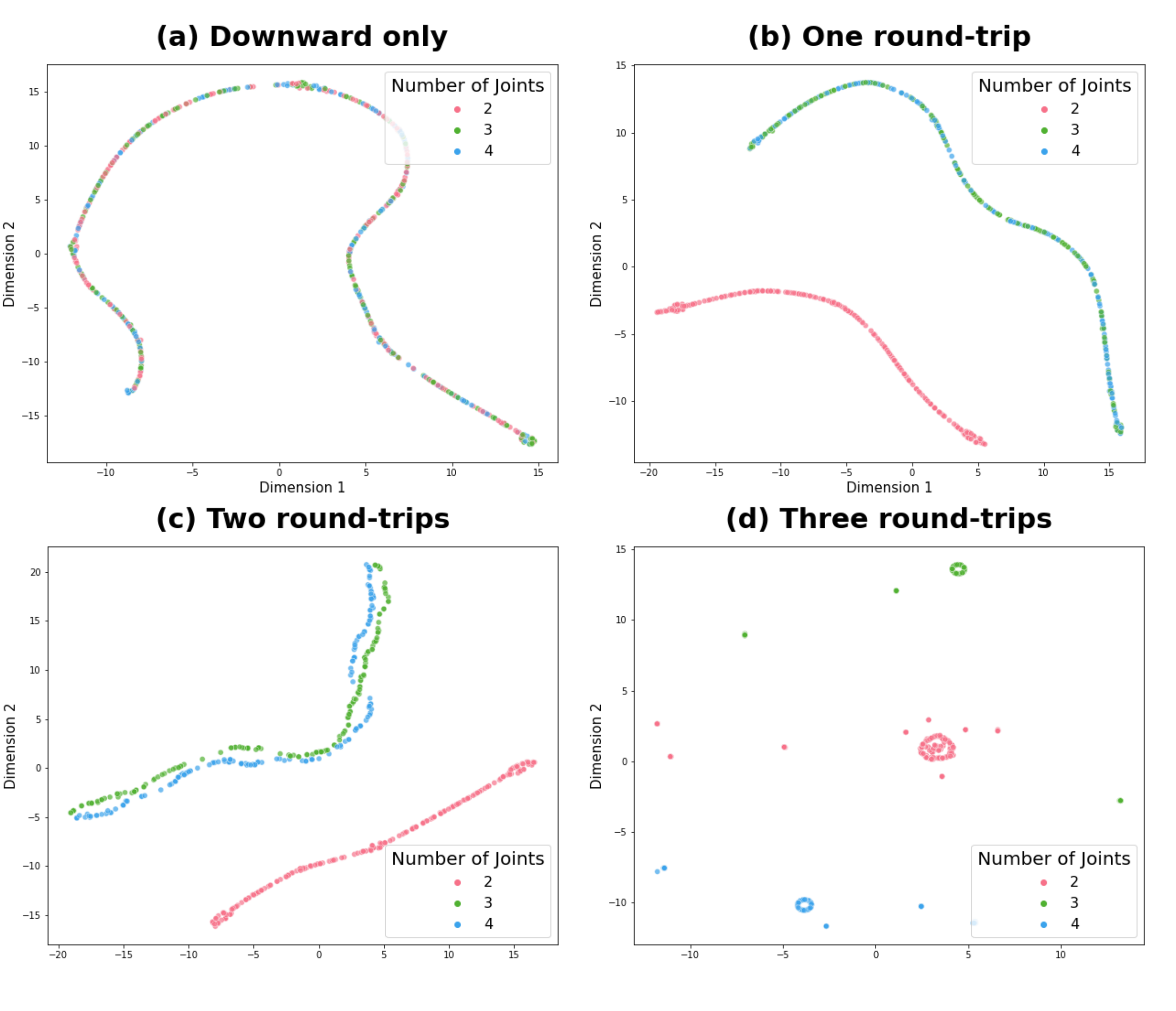}
    \vspace{-0.8cm}
    \caption{Comparison of structure embedding space learned from different message passing paths.
    (a) Downward only; (b) downward then upward, or one round-trip; (c) two round-trips; and (d) three round-trips.
    }
    \label{fig:passing-step}
\end{figure}

\subsection{Impacts of Message Passing on Embedding Space} \label{sec:exp_ablation}
We investigate how the message passing scheme affects the learned embedding through the lens of t-SNE visualizations.
Four different structure embedding spaces are visualized in Fig.~\ref{fig:passing-step}: (a) downward only; (b) downward then upward, or one round-trip; (c) two round-trips; and (d) three round-trips.
Different number of joints are depicted with different colors. 
With a single message passing (Fig.~\ref{fig:passing-step}a), the structures are likely to be sorted in a single line.
However, two-joint robots start to be separated from the cluster by having an additional pass from leaf to root.
Again, another round trip also managed to separate three- and four-joint structures.
However, adding one more round trip here shapes the embedding space differently. 
Instead of lines, circular clusters are formed per number of nodes, and additionally, multiple small groups are also spread around as if they are outliers.


%% file: 5_conclusion.tex
\section{Conclusion}
In this paper, we have proposed a tree-based message passing neural network for creating an embedding space for robotic structures and motions. 
We leverage the tree structure existing in kinematic structures to encode the given robot data into the corresponding embedding spaces.
We further incorporate the kinematic movements of the robot through the multi-task learning to learn more meaningful representations.
We visualized the learned embeddings via t-SNE dimensional reduction and analyzed how the embedding space formed for different structures and pose data.

There exist several different interesting future research directions. First, 
while we have tested embedding diverse structures in terms of joint number and link length, we only explored simple linear structures in this study. Therefore, we hope to add more complexity to the robot structure, such as more nodes, various types of revolute joints, and graph structures including closed chain.
This will also require investigating different aggregation strategies while accumulating the messages from multiple child nodes.
In addition, it would be interesting to extend this work by applying the learned embedding spaces to solve the tasks associated with robot structure, such as design optimization.

%% file: root.bbl
\begin{thebibliography}{10}
\providecommand{\url}[1]{#1}
\csname url@samestyle\endcsname
\providecommand{\newblock}{\relax}
\providecommand{\bibinfo}[2]{#2}
\providecommand{\BIBentrySTDinterwordspacing}{\spaceskip=0pt\relax}
\providecommand{\BIBentryALTinterwordstretchfactor}{4}
\providecommand{\BIBentryALTinterwordspacing}{\spaceskip=\fontdimen2\font plus
\BIBentryALTinterwordstretchfactor\fontdimen3\font minus
  \fontdimen4\font\relax}
\providecommand{\BIBforeignlanguage}[2]{{%
\expandafter\ifx\csname l@#1\endcsname\relax
\typeout{** WARNING: IEEEtran.bst: No hyphenation pattern has been}%
\typeout{** loaded for the language `#1'. Using the pattern for}%
\typeout{** the default language instead.}%
\else
\language=\csname l@#1\endcsname
\fi
#2}}
\providecommand{\BIBdecl}{\relax}
\BIBdecl

\bibitem{dong2017metapath2vec}
Y.~Dong, N.~V. Chawla, and A.~Swami, ``metapath2vec: Scalable representation
  learning for heterogeneous networks,'' in \emph{Proc. of the 23rd ACM SIGKDD
  international conference on knowledge discovery and data mining}, 2017, pp.
  135--144.

\bibitem{kolesnikov2019revisiting}
A.~Kolesnikov, X.~Zhai, and L.~Beyer, ``Revisiting self-supervised visual
  representation learning,'' in \emph{Proc. of the IEEE/CVF conference on
  computer vision and pattern recognition}, 2019, pp. 1920--1929.

\bibitem{chen2021exploring}
X.~Chen and K.~He, ``Exploring simple siamese representation learning,'' in
  \emph{Proc. of the IEEE/CVF Conference on Computer Vision and Pattern
  Recognition}, 2021, pp. 15\,750--15\,758.

\bibitem{gilmer2017neural}
J.~Gilmer, S.~S. Schoenholz, P.~F. Riley, O.~Vinyals, and G.~E. Dahl, ``Neural
  message passing for quantum chemistry,'' in \emph{Proc. of the International
  conference on machine learning}.\hskip 1em plus 0.5em minus 0.4em\relax PMLR,
  2017, pp. 1263--1272.

\bibitem{wu2018moleculenet}
Z.~Wu, B.~Ramsundar, E.~N. Feinberg, J.~Gomes, C.~Geniesse, A.~S. Pappu,
  K.~Leswing, and V.~Pande, ``Moleculenet: a benchmark for molecular machine
  learning,'' \emph{Chemical science}, vol.~9, no.~2, pp. 513--530, 2018.

\bibitem{duvenaud2015convolutional}
D.~Duvenaud, D.~Maclaurin, J.~Aguilera-Iparraguirre, R.~G{\'o}mez-Bombarelli,
  T.~Hirzel, A.~Aspuru-Guzik, and R.~P. Adams, ``Convolutional networks on
  graphs for learning molecular fingerprints,'' \emph{arXiv preprint
  arXiv:1509.09292}, 2015.

\bibitem{mansimov2019molecular}
E.~Mansimov, O.~Mahmood, S.~Kang, and K.~Cho, ``Molecular geometry prediction
  using a deep generative graph neural network,'' \emph{Scientific reports},
  vol.~9, no.~1, pp. 1--13, 2019.

\bibitem{reily2020representing}
B.~Reily, C.~Reardon, and H.~Zhang, ``Representing multi-robot structure
  through multimodal graph embedding for the selection of robot teams,'' in
  \emph{Proc. of the International Conference on Robotics and
  Automation}.\hskip 1em plus 0.5em minus 0.4em\relax IEEE, 2020, pp.
  5576--5582.

\bibitem{ha2017joint}
S.~Ha, S.~Coros, A.~Alspach, J.~Kim, and K.~Yamane, ``Joint optimization of
  robot design and motion parameters using the implicit function theorem.'' in
  \emph{Robotics: Science and systems}, vol.~8, 2017.

\bibitem{ha2018computational}
------, ``Computational co-optimization of design parameters and motion
  trajectories for robotic systems,'' \emph{The International Journal of
  Robotics Research}, vol.~37, no. 13-14, pp. 1521--1536, 2018.

\bibitem{zhao2020robogrammar}
A.~Zhao, J.~Xu, M.~Konakovi{\'c}-Lukovi{\'c}, J.~Hughes, A.~Spielberg, D.~Rus,
  and W.~Matusik, ``Robogrammar: graph grammar for terrain-optimized robot
  design,'' \emph{ACM Transactions on Graphics (TOG)}, vol.~39, no.~6, pp.
  1--16, 2020.

\bibitem{xu2021multi}
J.~Xu, A.~Spielberg, A.~Zhao, D.~Rus, and W.~Matusik, ``Multi-objective graph
  heuristic search for terrestrial robot design,'' \emph{arXiv preprint
  arXiv:2107.05858}, 2021.

\bibitem{wang2019neural}
T.~Wang, Y.~Zhou, S.~Fidler, and J.~Ba, ``Neural graph evolution: Towards
  efficient automatic robot design,'' \emph{arXiv preprint arXiv:1906.05370},
  2019.

\bibitem{wang2018nervenet}
T.~Wang, R.~Liao, J.~Ba, and S.~Fidler, ``Nervenet: Learning structured policy
  with graph neural networks,'' in \emph{Proc. of the International Conference
  on Learning Representations}, 2018.

\bibitem{huang2020one}
W.~Huang, I.~Mordatch, and D.~Pathak, ``One policy to control them all: Shared
  modular policies for agent-agnostic control,'' in \emph{Proc. of the
  International Conference on Learning RepresentationsInternational Conference
  on Machine Learning}.\hskip 1em plus 0.5em minus 0.4em\relax PMLR, 2020, pp.
  4455--4464.

\bibitem{whitman2021learning}
J.~Whitman, M.~Travers, and H.~Choset, ``Learning modular robot control
  policies,'' \emph{arXiv preprint arXiv:2105.10049}, 2021.

\bibitem{zhang2014facial}
Z.~Zhang, P.~Luo, C.~C. Loy, and X.~Tang, ``Facial landmark detection by deep
  multi-task learning,'' in \emph{Proc. of the European conference on computer
  vision}.\hskip 1em plus 0.5em minus 0.4em\relax Springer, 2014, pp. 94--108.

\bibitem{dai2016instance}
J.~Dai, K.~He, and J.~Sun, ``Instance-aware semantic segmentation via
  multi-task network cascades,'' in \emph{Proc. of the IEEE conference on
  computer vision and pattern recognition}, 2016, pp. 3150--3158.

\bibitem{xu2018pad}
D.~Xu, W.~Ouyang, X.~Wang, and N.~Sebe, ``Pad-net: Multi-tasks guided
  prediction-and-distillation network for simultaneous depth estimation and
  scene parsing,'' in \emph{Proc. of the IEEE Conference on Computer Vision and
  Pattern Recognition}, 2018, pp. 675--684.

\bibitem{vandenhende2020mti}
S.~Vandenhende, S.~Georgoulis, and L.~Van~Gool, ``Mti-net: Multi-scale task
  interaction networks for multi-task learning,'' in \emph{Proc. of the
  European Conference on Computer Vision}.\hskip 1em plus 0.5em minus
  0.4em\relax Springer, 2020, pp. 527--543.

\bibitem{liebel2018auxiliary}
L.~Liebel and M.~K{\"o}rner, ``Auxiliary tasks in multi-task learning,''
  \emph{arXiv preprint arXiv:1805.06334}, 2018.

\bibitem{giannone2019learning}
G.~Giannone and B.~Chidlovskii, ``Learning common representation from rgb and
  depth images,'' in \emph{Proc. of the IEEE/CVF Conference on Computer Vision
  and Pattern Recognition Workshops}, 2019, pp. 0--0.

\bibitem{yu2015visual}
L.~Yu, E.~Park, A.~C. Berg, and T.~L. Berg, ``Visual madlibs: Fill in the blank
  description generation and question answering,'' in \emph{Proceedings of the
  ieee international conference on computer vision}, 2015, pp. 2461--2469.

\bibitem{joshi2020spanbert}
M.~Joshi, D.~Chen, Y.~Liu, D.~S. Weld, L.~Zettlemoyer, and O.~Levy, ``Spanbert:
  Improving pre-training by representing and predicting spans,''
  \emph{Transactions of the Association for Computational Linguistics}, vol.~8,
  pp. 64--77, 2020.

\bibitem{devlin2018bert}
J.~Devlin, M.-W. Chang, K.~Lee, and K.~Toutanova, ``Bert: Pre-training of deep
  bidirectional transformers for language understanding,'' \emph{arXiv preprint
  arXiv:1810.04805}, 2018.

\bibitem{bertalmio2000image}
M.~Bertalmio, G.~Sapiro, V.~Caselles, and C.~Ballester, ``Image inpainting,''
  in \emph{Proceedings of the 27th annual conference on Computer graphics and
  interactive techniques}, 2000, pp. 417--424.

\bibitem{brockman2016openai}
G.~Brockman, V.~Cheung, L.~Pettersson, J.~Schneider, J.~Schulman, J.~Tang, and
  W.~Zaremba, ``Openai gym,'' \emph{CoRR}, vol. abs/1606.01540, 2016.

\bibitem{coumans2019pybullet}
E.~Coumans and Y.~Bai, ``Pybullet, a python module for physics simulation for
  games, robotics and machine learning,'' \url{http://pybullet.org},
  2016--2019.

\bibitem{kingma2014adam}
D.~P. Kingma and J.~Ba, ``Adam: A method for stochastic optimization,''
  \emph{arXiv preprint arXiv:1412.6980}, 2014.

\bibitem{van2008visualizing}
L.~Van~der Maaten and G.~Hinton, ``Visualizing data using t-sne.''
  \emph{Journal of machine learning research}, vol.~9, no.~11, 2008.

\end{thebibliography}
